\documentclass[review]{elsarticle}

\usepackage{lineno,hyperref}
\modulolinenumbers[5]

\journal{Engineering Applications of Artificial Intelligence}
\usepackage{amsmath}
\usepackage{graphics}
\usepackage{graphicx}
\usepackage{float}
\usepackage[T1]{fontenc}
\usepackage[utf8]{inputenc}
\usepackage{tabularx}
\usepackage{balance}
\usepackage{ amssymb }
\usepackage{mathtools}
\usepackage{gensymb}
\usepackage{subfig}
\usepackage{multicol}
\usepackage{lipsum}
\usepackage{chngpage}
\usepackage{calc}
\usepackage{ragged2e}
\usepackage{tabularx}
\usepackage[margin=2.5cm]{geometry}
\usepackage{booktabs}
\usepackage{algorithm}
\usepackage[noend]{algorithmic}

\newcommand\algorithmicinput{\textbf{INPUT}}
\newcommand\INPUT{\item[\algorithmicinput]}
\renewcommand{\algorithmicinput}{\textbf{ClientUpdate}}

\makeatletter
\def\ps@pprintTitle{%
  \let\@oddhead\@empty
  \let\@evenhead\@empty
  \def\@oddfoot{\reset@font\hfil\thepage\hfil}
  \let\@evenfoot\@oddfoot
}
\makeatother

\begin{document}

\begin{frontmatter}

\title{FedTrees: A Novel Computation-Communication Efficient Federated Learning Framework Investigated in Smart Grids}


\author[1]{~Mohammad~Al-Quraan\corref{mycorrespondingauthor}}
\author[1]{~Ahsan~Khan}
\author[1]{~Anthony~Centeno}
\author[1]{~Ahmed~Zoha}
\author[1]{~Muhammad~Ali~Imran}
\author[1]{ and Lina~Mohjazi}

\address[1]{ James Watt School of Engineering,
University of Glasgow, Glasgow, G12 8QQ, UK, \\(e-mail: \{m.alquraan.1, a.khan.9\}@research.gla.ac.uk,\\ \{Anthony.Centeno, Ahmed.Zoha, Muhammad.Imran, Lina.Mohjazi\}@glasgow.ac.uk).} 

\cortext[mycorrespondingauthor]{Corresponding author}

\begin{abstract}
Smart energy performance monitoring and optimisation at the supplier and consumer levels is essential to realising smart cities. In order to implement a more sustainable energy management plan, it is crucial to conduct a better energy forecast. The next-generation smart meters can also be used to measure, record, and report energy consumption data, which can be used to train machine learning (ML) models for predicting energy needs. However, sharing fine-grained energy data and performing centralised learning may compromise users' privacy and leave them vulnerable to several attacks. This study addresses this issue by utilising federated learning (FL), an emerging technique that performs ML model training at the user level, where data resides. We introduce FedTrees, a new, lightweight FL framework that benefits from the outstanding features of ensemble learning. Furthermore, we developed a delta-based early stopping algorithm to monitor FL training and stop it when it does not need to continue. The simulation results demonstrate that FedTrees outperforms the most popular federated averaging (FedAvg) framework and the baseline Persistence model for providing accurate energy forecasting patterns while taking only 2\% of the computation time and 13\% of the communication rounds compared to FedAvg, saving considerable amounts of computation and communication resources.
\end{abstract}

\end{frontmatter}


\section{Introduction}
Fifth-generation (5G) and beyond wireless technologies are expected to unlock the full potential of the Internet of Things (IoT), the key enabler of the smart city model \cite{[25]}. The smart city concept combines several elements such as a smart environment, mobility, living, and energy to improve citizens’ quality of life. Fulfilling smart energy, in the form of smart grids and smart buildings, has been the focus of many bodies in industry and academia through monitoring and predicting energy consumption patterns \cite{[26]}. Accurate short-, mid-, or long-term energy forecasting is the ultimate goal that helps managers or consumers to prepare better future plans, thus improving energy performance.

Energy suppliers need to maintain an equilibrium point between supply and demand, since producing excessive amounts of energy will result in energy wastage. In contrast, failure to meet consumers' demands may lead to the need to purchase energy at higher rates; otherwise, frequent blackouts will happen. Therefore, various load forecasting techniques have been considered for electricity networks' efficient and reliable operation. Statistical forecasting methods, e.g., multiple linear regression (MLR), autoregressive (AR), and moving average (MA) techniques, were used to project past and present load profiles into future predictions. Later, the introduction of smart metering and the evolution of artificial intelligence (AI) technology paved the way for replacing traditional prediction techniques with various machine learning (ML) algorithms, due to their ability in analysing large amounts of datasets in short periods of time while providing impressive accuracy levels \cite{[27]}. Advanced metering infrastructure (AMI), a system of smart meters connected to a communication network for two-way communications between customers and utility companies, is the first step toward smart energy, which helps collect and analyse smart-meter data. However, collecting consumers' load profiles into a central entity to conduct energy forecasting raises privacy concerns. Individuals' load information could be misused by revealing consumer habits and household occupancy.

To address this issue, the ML community recently introduced a new learning paradigm termed as federated learning (FL) \cite{[28]}. The FL is analogous to the concept of distributed learning in terms of handling enormous datasets and developing efficient and scalable systems. However, maintaining data privacy is the goal of FL as it does not involve the collection of data in a central location, but instead it sends the model to the clients where the data is generated. The FL framework is orchestrated by a server placed in a central entity, i.e., an energy supplier, to train and improve a shared model with many clients collaboratively. Two typical FL architectures exist based on the scale of the federation. The first is cross-device, where the number of clients may be massive, for example, consumers’ smart meters. The second is cross-silo, which considers relatively limited and reliable clients, for example, substations. The FL process starts by initialising a global model in the server and then sending it to the clients to conduct model training. Once completed, the clients send back the model updates to the server, which will aggregate them, resulting in an updated model. Then, the updated model is sent to the clients for another training round. This process is repeated until the limit of communication rounds is reached, or the model achieves the desired accuracy.

The use of FL in energy forecasting is still in very early stages, and few studies have considered this approach \cite{[20],[21],[22],[23],[24]}. In these studies, the authors focus on utilising long short-term memory (LSTM) architectures, a type of recurrent neural network (RNN) used in the field of deep learning (DL), due to their remarkable performance in predicting time-series data sequences. However, the mentioned works overlook a critical issue: DL models are extremely resource-consuming (energy, memory, processor, etc.), and the lengthy and extensive underlying mathematical operations demand resource-rich hardware. Considering such schemes of combining FL with LSTM models requires extended computation time to reach the desired precision and impede their scalability. Furthermore, individual households are the focal point of the abovementioned studies to be used as FL clients. Conversely, our study applies FL at the substation level, i.e., a part of the power system dedicated to servicing local dwellings in a specific area, allowing for the use of FL algorithm within one or more energy suppliers, hence attaining more generalised ML models. 

In this paper, we propose FedTrees; a novel light aggregation algorithm developed to utilise decision trees (DTs) under the FL setting. Specifically, we use the light gradient boosting model (LGBM) \cite{[29]}, one of the boosting techniques in ensemble learning, to be sent and trained across the clients of the FL framework. The main reasons for considering the LGBM models are due to their rapid training speed, less memory consumption, higher efficiency, and accurate predictions. FedTrees attempts to minimise computations and the number of communication rounds while guaranteeing high training performance; hence it is envisioned to play a crucial role in a wide range of FL-based applications in several fields, such as smart energy. Moreover, this work considers the common challenge of optimising the number of communication rounds that may lead to suboptimal performance or excessive rounds of unnecessary training, thus consuming computation and communication resources. Therefore, we developed a delta-based early stopping patience technique, a dynamic algorithm that monitors the FL training process and stops it when no further enhancement is possible. Also, this paper examines the importance of each feature used in the training process and offers a study of the effect of using a different number of features on the final training performance. Additionally, the performance of the FedTrees algorithm is benchmarked against the popular LSTM-based federated averaging (FedAvg) and the naïve Persistence model. Finally, the proposed framework is evaluated based on state-of-the-art metrics and by conducting extensive simulations. The main contributions of the paper are summarised as follows:

\begin{itemize}
    \item First, we introduce FedTrees, a novel, light algorithm that employs DT-based models within the FL setup. FedTrees using LGBM models shows improved performance in terms of the required number of communication rounds and computation time compared to LSTM-based FedAvg aggregation scheme when employed for energy forecasting.

    \item Unlike other FL-based energy forecasting framework that rely on fixing the number of communication rounds, this study develops a delta-based early stopping technique to reach the best possible accuracy while reducing the computation and communications costs.
    
    \item A feature importance evaluation study is conducted against individual load profiles for optimal forecasting performance.
    
    \item Finally, we compared FedTrees performance with LSTM-based FedAvg and a naïve Persistence model as a benchmark using state-of-the-art metrics. The results reveal a significant improvement in overall performance using the FedTrees framework.

\end{itemize}

The remainder of this paper is structured as follows. Section \ref{RW} recalls the related work, while Section \ref{BG} presents some background notions. Section \ref{Framework} describes the proposed FedTrees algorithm versus FedAvg and represents the delta-based early stopping patience technique. Simulation setup, used dataset, and numerical results are presented in Section \ref{PE}. Finally, the concluding remarks are drawn in Section \ref{Conclusion}.

\section{Related Work}\label{RW}
This section first reviews the works that use different load forecasting methodologies performed either locally or centrally; then, it presents the state-of-the-art research that takes into account the FL framework for load forecasting.

\subsection{Local and Centralised Approaches}
Load profiles store energy/power consumption information in the form of time-series data, which can be predicted using several approaches, such as statistical and computational intelligence. For a complete view of the energy forecasting methods, we refer the reader to \cite{[0]}. Statistical methods have been used in the literature for the short and medium temporal forecasting ranges and show quite good performance. For instance, the study in \cite{[1]} uses the MLR technique to verify its reliability in energy demand forecasting instead of traditional and more complex methods. The authors in \cite{[3]} rely on the traditional AR methodology to predict electrical energy in Lebanon. Using univariate historical time series data, the second-order MA model is combined with another statistical method for forecasting gas consumption in China from 2009 to 2015 \cite{[4]}. Moreover, combining AR and MA resulted in many variants that enhance the forecasting process such as the autoregressive moving average (ARMA) \cite{[5]}, ARMA model with exogenous inputs (ARMAX) \cite{[6]}, AR integrated MA (ARIMA) model \cite{[2]}, and seasonal ARIMA (SARIMA) \cite{[7]}.

Later, the focus on ML methods became dominant owing to the advantages of AI in analysing large amounts of data, which was made available by the introduction of smart metering. The contribution in \cite{[9]} highlights the use of support vector machine (SVM) to forecast the monthly electrical load of Taiwan. On the other hand, many studies use NNs in the energy forecasting domain by virtue of their impressive performance in various areas. In \cite{[10]}, the work uses electricity time-series data from three countries to compare the performance of convolutional NN (CNN) and multilayer perceptron (MLP) algorithms. Similarly, the authors in \cite{[11]} demonstrate that a three hidden layer CNN model performs well in energy forecasting compared to statistical and other simple ML methods. RNN is also widely used in smart energy, particularly LSTM and its variants \cite{[12],[13]}. 

Apart from NNs, DTs have also been used in the energy forecasting task. Although the DT approach is simple, it shows desired performance when predicting future energy consumption in Hong Kong \cite{[15]}. Following the same DT approach, the authors in \cite{[16]} demonstrate that employing DT for predicting energy use intensity in Japanese residential buildings can give very well accuracy. However, DT alone has not been widely used because it suffers from several drawbacks, such as instability, easily losing generalisation, and performing poorly with noisy and non-linear data. Later, ensemble-based algorithms gained popularity and were explored in the energy research domain. In general, ensemble techniques possess attractive features that draw the research community's attention, such as simplicity, ease of use, interpretability, and computational efficiency. For instance, random forest forecasting performance is compared with NN in \cite{[17]}, and the results demonstrated that both are feasible and effective in building energy applications. The gradient-boosted decision tree (GBDT) and extreme gradient boosting (XGBoost) algorithms are also utilised in predicting future electricity loads and have shown to be effective in \cite{[18]} and \cite{[19]}, respectively.

The proposed studies and frameworks mentioned above are generally based on a centralised model training where the energy consumption information is transmitted across the network and combined in a central location. However, this training scheme raises privacy concerns. Consumers’ load profiles hold sensitive information that can be used in various dimensions like inferring occupancy and usage patterns of households, government surveillance, data selling, and illegal data use among others.

\subsection{FL-based Load Forecasting}
Recently, the ML community has investigated a new research direction to address the privacy concerns associated with traditional training methods. As a distributed learning algorithm, FL can perform model training at the edge of the network rather than a central location without the need to share sensitive load information. Few studies have begun to consider the use of FL in the context of smart energy; for example, the aim of the work in \cite{[20]} is to evaluate the use of federated settings in predicting electrical load consumption patterns that will assist in load monitoring and energy demand response. The FedAvg technique is applied to aggregate the parameters of LSTM models while conducting training, and to produce a generalised global model. Applying this framework to several households in the USA, the study demonstrated that FL has a great potential in smart grids through training a powerful future energy forecasting model. Similarly, FedAvg and LSTM are adopted in \cite{[21]} to provide a generalised electrical load forecasting model. The authors use complementary features related to calendar and weather conditions along with the sequences of previous electrical load to improve the forecast model.

Briggs \emph{et al.} \cite{[22]} present a study on the importance of using smart meters in residential areas to forecast energy consumption using FL, which helps move towards renewable energy generation. LSTM algorithm is considered to perform short-term forecasting tasks. Similarly, the contribution in \cite{[23]} evaluates the performance of FL versus centralised and local training methods when using LSTM models in the context of electrical load forecasting. In addition, clustering is considered to group the clients based on similar model hyperparameters and, accordingly, similar data characteristics. The study concluded that local learning is better for predicting individual energy consumption than FL. However, FL is needed when a generalised forecasting model is required and access to aggregated data is impossible. Very recently, the research work in \cite{[24]} adopted the LSTM algorithm under the FL setting to forecast energy profiles. Two strategies are considered, namely, federated stochastic gradient descent (FedSGD) and FedAvg to perform models’ parameter aggregation. Experimental results demonstrated that FedAvg achieves better accuracy and requires fewer communication rounds.

The aforementioned FL-based energy forecasting studies rely on DL algorithms, specifically LSTM networks. Although the LSTMs have shown to achieve excellent prediction accuracy, they require intensive processing duties that yield a heavy computational burden. The following section provides a background to the main elements considered for conducting this study.

\section{Preliminaries and background}\label{BG}
In this section, we review the basic concepts of LSTM, LGBM, and FL needed to understand better the study conducted in this paper.

\subsection{Long Short-Term Memory (LSTM) Networks}
\begin{figure}
\centering
\includegraphics[scale=0.38]{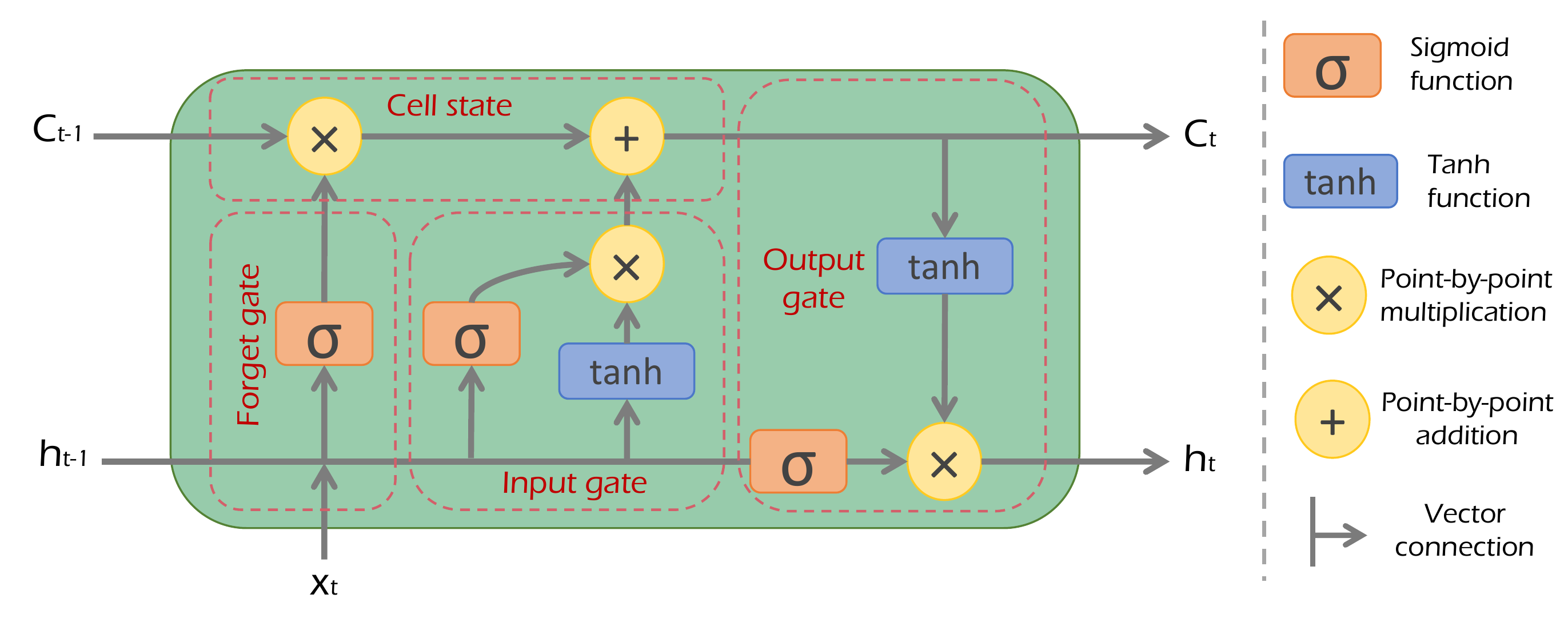}
\caption{LSTM memory cell with gating units.}
\label{LSTM}
\end{figure}

An RNN is a class of artificial neural networks designed to process sequential data through which it recognises patterns, understands temporal dynamic behaviours, and provides predictions based on the previous states \cite{[38]}. As in NNs, RNNs are gradient-based learning algorithms that rely on the backpropagation mechanism to update neurons' weights. However, RNNs suffer from two main challenges associated with performing partial derivatives across the network to find the model weights: vanishing and exploding gradient. These problems may prevent the network from further training, causing the network to struggle to learn long-term dependencies. Therefore, LSTM networks were introduced as more robust models without being affected by the unstable gradients problem by improving the gradient flow \cite{[38]}. This was done by introducing artificial memory (cell state) and three gates: a forget gate, an input gate, and an output gate. The three gates can be thought of as filters that regulate the flow of information in and out of the cell to help predict the output sequence. Specifically, the forget gate decides which information to keep/neglect from the previous cell state. The input gate decides which information is relevant to update the cell state. On the other hand, the output gate determines the new hidden state of the next LSTM unit. A pictorial representation of an LSTM unit is given in Fig. \ref{LSTM}.

\subsection{Light Gradient Boosting Model (LGBM)}
DTs are supervised ML algorithms that can do regression and classification by continually splitting data according to specific rules. The simplicity of DTs has made them popular, and they have been applied in many applications. However, DTs suffer from several challenges that limit their use in more complex situations, such as overfitting, instability, and bias. Therefore, the concept of ensemble learning was adopted with the aim of combining many weaker learners (i.e., DTs) to produce a more robust ensemble. The ensemble learning includes three main classes, bagging, stacking, and boosting. For more details on ensemble learning, we refer the reader to \cite{[30]}.

\begin{figure}
\centering
\includegraphics[scale=0.5]{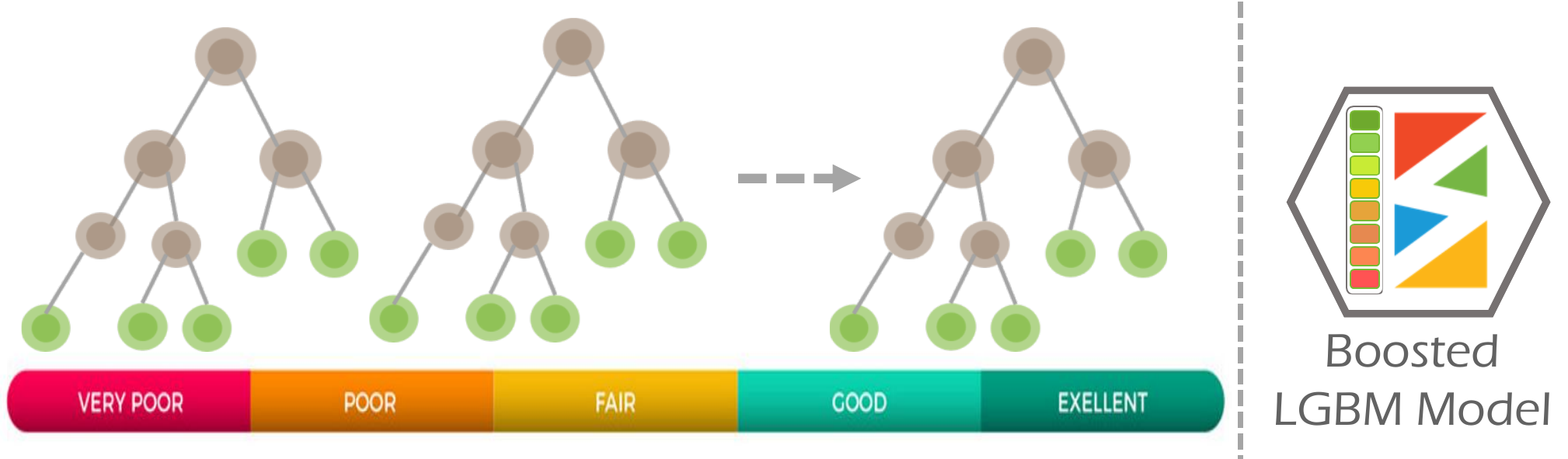}
\caption{Ensemble of DTs are combined to boost and form the LGBM model.}
\label{LGBM}
\end{figure}

In 2017, Microsoft introduced one of the most powerful boosting models named LGBM \cite{[29]}, depicted in Fig. \ref{LGBM}. What distinguishes LGBM from other boosting models is its efficiency, fast processing, and scalability. These desirable features are gained by introducing two techniques: gradient-based one-side sampling (GOSS) and exclusive feature bundling (EFB). GOSS downsample data instances with small absolute gradients while keeping the samples with higher absolute gradients because they contribute more to the training process. At the same time, EFB reduces the number of data features by bundling the mutually exclusive ones. Moreover, unlike other tree-based algorithms that grow trees horizontally, i.e., level-wise, LGBM grows trees vertically, i.e., leaf-wise, by choosing the leaf that it believes will achieve the highest loss decrease.  It is worth mentioning that the level-wise strategy is better for smaller datasets to avoid overfitting, while the leaf-wise strategy excels in larger datasets. Lower computations, less memory consumption, fast training, handling large-scale data, and scalability are the main characteristics that will help the prevalence of LGBM algorithms in solving many real-world problems.

\subsection{Federated Averaging}
ML is a data-driven approach that relies on several quantities of data sets to train a model for a specific task. The conventional way to conduct model training is by collecting the required datasets at a central location, performing pre-processing, and then feeding them to the model for training. However, centralised training methods threaten data privacy and security and contradict the legislation in data protection laws. Therefore, FL has emerged as a promising solution that addresses privacy concerns by pushing the model to the locations where the data is generated and exists. The FL framework is an iterative training process that incorporates specific steps to achieve the desired performance. In short, the FL training procedure includes initiating and broadcasting the global model to the clients, performing local model training, uploading model updates to the server, and finally aggregating the updates to produce a new version of the model \cite{[28]}.

Several aggregation algorithms have been introduced such as FedAvg \cite{[31]}, FedProx \cite{[32]}, and FedDist \cite{[33]}. However, these algorithms are designed to aggregate the parameters of NNs models and cannot be used with other ML algorithms, for example, SVMs and DTs. Furthermore, the FL framework is highly dependent on the resources of clients and networks, and using NN models under the FL framework will lead to highly consuming these resources. Therefore, building new efficient aggregation schemes is crucial. To tackle this problem, this study proposes a novel tree based FL framework, termed as FedTrees, which will be discussed in detail in the next section. FedTrees performance is compared with FedAvg, the most widely used aggregation algorithm in the literature. 

Finding the minimum of the loss function is fundamental in evaluating the performance of the ML algorithm in dataset modelling. Various ML algorithms depend on gradient descent (GD). This optimisation algorithm attempts to track down the minimum value of a differential function by finding the gradient at a given point and then taking steps proportional to the negative of the gradient. However, this algorithm demands a considerable computation complexity proportional to the size of the dataset. Then, a stochastic GD (SGD) is introduced to relieve the computations by performing GD on batches of data samples. SGD has improved the model convergence rate and is used under the FL setting (FedSGD) as the first naïve aggregation algorithm. Although FedSGD eases the computation complexity on the client-side, it requires many communication rounds with the central server. Therefore, the FedAvg algorithm is developed to alleviate the pressure on communication resources by doing multiple minibatch gradient calculations before updating the server. The amount of computation is controlled using three parameters; a subset of total client count (C), breaking down the dataset into small-sized mini-batches (B), and the number of epochs (E) that the client passes over its dataset per round. FedAvg allows each client to perform multiple SGD rounds locally across different local data subsets and then find the optimal model parameters by averaging the clients’ locally evaluated gradients at the FL server. The complete pseudo-code of FedAvg is given in Algorithm \ref{alg1}.

FedAvg is designed to best fit neural networks where network parameters (weights and biases) can be extracted and aggregated to form newly updated parameters. LSTM networks, one form of neural networks, have been widely used in predicting time-series datasets, and they fit the federated optimisation problem. In previous studies that considered energy forecasting tasks under the FL environment, their primary model choice was certainly the LSTM networks. Although the prediction performance of LSTM models is accurate, these studies have overlooked very crucial problems accompanied by FL distributed learning, which are the computation and communication costs.

\begin{algorithm}[t]
\caption{Federated Averaging \cite{[31]}}\label{alg1}
\begin{algorithmic}
\REQUIRE Communications limit ($T$), Clients count ($C$), Number of mini-batches ($B$), Number of epochs ($E$)\\
\ENSURE
\STATE $w_{0}\leftarrow$ initialise weights
    \FOR{$t = 1,2,...,T$}
    \STATE $S_{t} = $ Random set of $C$
        \FOR{$k = 1,2,...,S_{t}$ \textbf{in parallel}}
        \STATE $w_{t}^{k}\leftarrow$ ClientUpdate($k,w_{t-1}$)
        \ENDFOR
    \STATE $w_{t}\leftarrow \sum_{k=1}^{S_{t}} \frac {n_{k}}{n} w_{t}^{k} $
    \ENDFOR
\STATE
\INPUT($k,w$)\textbf{:}
    \FOR{$e = 1,2,...,E$}
        \STATE $batches \leftarrow$ Split dataset into $B$ batches
        \FOR{$b = 1,2,...,batches$}
            \STATE $w \leftarrow w-\eta \nabla l(w;b)$ 
            \begin{footnotesize}
           //\textit{Compute gradients and update weights}
            \end{footnotesize}
        \ENDFOR
    \ENDFOR
\STATE Return $w$ to server
\end{algorithmic}
\end{algorithm}

\section{Proposed FedTrees Algorithm} \label{Framework}
 
\begin{figure*}
\centering
\includegraphics[scale=0.52]{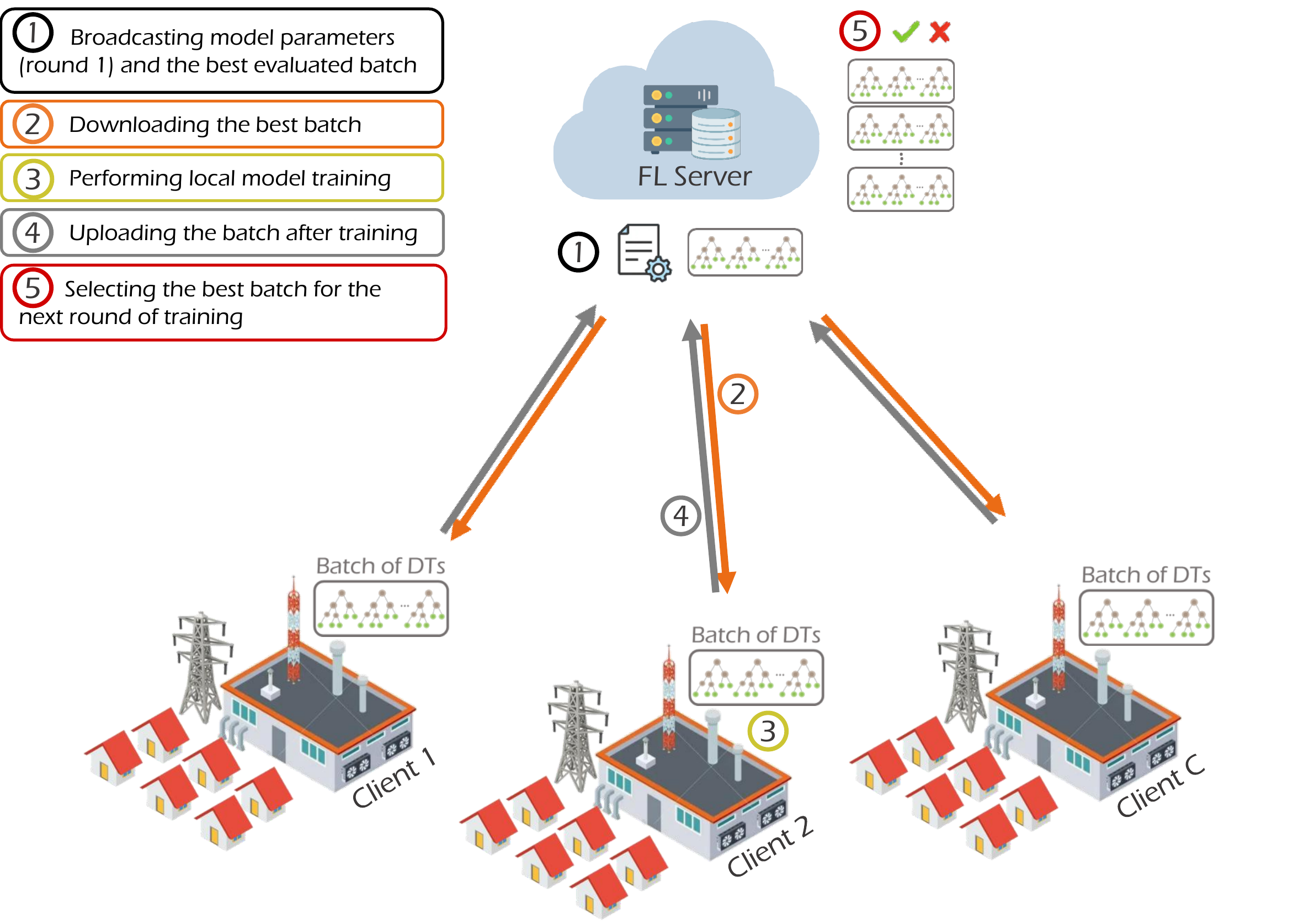}
\caption{FedTrees sequential operation steps for energy forecasting task by considering C substations as clients.}
\label{FedTrees}
\end{figure*}

\begin{algorithm}[t]
\caption{FedTrees}\label{alg2}
\begin{algorithmic}
\REQUIRE Communications rounds ($R$), Clients count ($C$), Number of trees in one batch ($T$)\\
\ENSURE
    \FOR{$r = 1,2,...,R$}
    \IF {$r == 1$}
    \STATE Broadcast LGBM parameters values to clients $Batch^{0} = \{\}$
    \ENDIF
    \STATE Broadcast $Batch^{r-1}$
        \FOR{$k = 1,2,...,C$ \textbf{in parallel}}
        \STATE Store $Batch^{r-1}$ in memory 
        \vspace{0.1 cm}
        \STATE $Batch_{k}^{r}\leftarrow$ ClientUpdate($k,\sum_{i=0}^{r-1}{Batch^{i}},r$)
        \vspace{0.1 cm}
        \ENDFOR
         \begin{small}
           //\textit{Find the best Batch}
        \end{small}
        \STATE $Batch^{r}$ $\leftarrow$ \small {Server validation} $\{\sum_{i=0}^{r-1}{Batch^{i}+Batch_{k}^{r}}\}_{k=1}^{C}$
    \ENDFOR
\STATE
\INPUT($k, \sum_{i=0}^{r-1}{Batch^{i}}, r$)\textbf{:}
\vspace{0.1 cm}
\STATE $Model_{k}^{r} \leftarrow \sum_{i=0}^{r-1}{Batch^{i}} +\sum_{t=1}^{T}{DT_{t}}$
\vspace{0.1 cm}
    \FOR{$t = 1,2,...,T$}
        \STATE Find the optimal split for each split node in the new batch trees and update $Model_{k}^{r}$
    \ENDFOR
\vspace{0.1 cm}
\STATE Return $\sum_{t=1}^{T}{DT_{t}}$
\end{algorithmic}
\end{algorithm}

\subsection{FedTrees in LGBM-based FL}
As mentioned earlier, in the federal environment, the use of NNs is becoming more and more popular. However, many other ML algorithms have not had the opportunity to be explored in such an environment, even though they conceal many wanted merits like simple but efficient approaches. Recently, the research community has begun to investigate the use of other ML techniques, such as DTs \cite{[34]}, \cite{[35]}. This study follows the same concept of employing DTs under the FL setting; however, it is the first research that explores the use of LGBM models with FL. This study is also the leading in exploring LGBM-based FL in the context of smart grids. Inspired by the FedVoting algorithm presented in \cite{[35]}, where the authors construct a federated GBDT model trained for human mobility prediction, this study proposes FedTrees. FedTrees framework harnesses the power of the LGBM algorithm within the FL environment. What distinguishes FedTrees from FedVoting is that it is less complex and scalable. To be precise, FedVoting is developed for the cross-silo setup and relies on cross-validations. To determine the optimal model for each training round, each client must validate other clients’ trained local models, incurring additional computation cost. Moreover, scalability is not by design of FedVoting since it is designed to be performed on a limited number of clients. By contrast, FedTrees is developed to fulfil simplicity, efficiency, and scalability, while fitting both cross-device and cross-silo settings. In FedTrees, the complexity is alleviated as there is no need for any client to validate others’ models; the central server validates the received models and selects the best one to build upon in upcoming training rounds.

Fig. \ref{FedTrees} demonstrates the structure of FedTrees and the detailed training process.  Constructing FedTrees begins with choosing the best hyperparameters of the LGBM model, like the number of estimators (trees), the number of leaves for each estimator, boosting type, and max depth. Once determined, the central server broadcasts these parameters to each participating client except for the number of estimators, which is the server's role in determining the best initial batch of trees to be trained in each round and build on it. When every client receives the model parameters, they create the LGBM model accordingly and start the iterative model training. At the end of each training round, the clients send the trained batch of trees to the central server for evaluation. The batch that achieves the best evaluation performance will be picked to build on in the following communication rounds, while the other batches will be discarded. This process is repeated until the global LGBM model reaches the desired accuracy or no further improvement can be achieved. The complete FedTrees pseudo-code is presented in Algorithm \ref{alg2}.

\subsection{Delta-based Early Stopping Patience} \label{Stopping}
\begin{figure}
\centering
\includegraphics[scale=0.8]{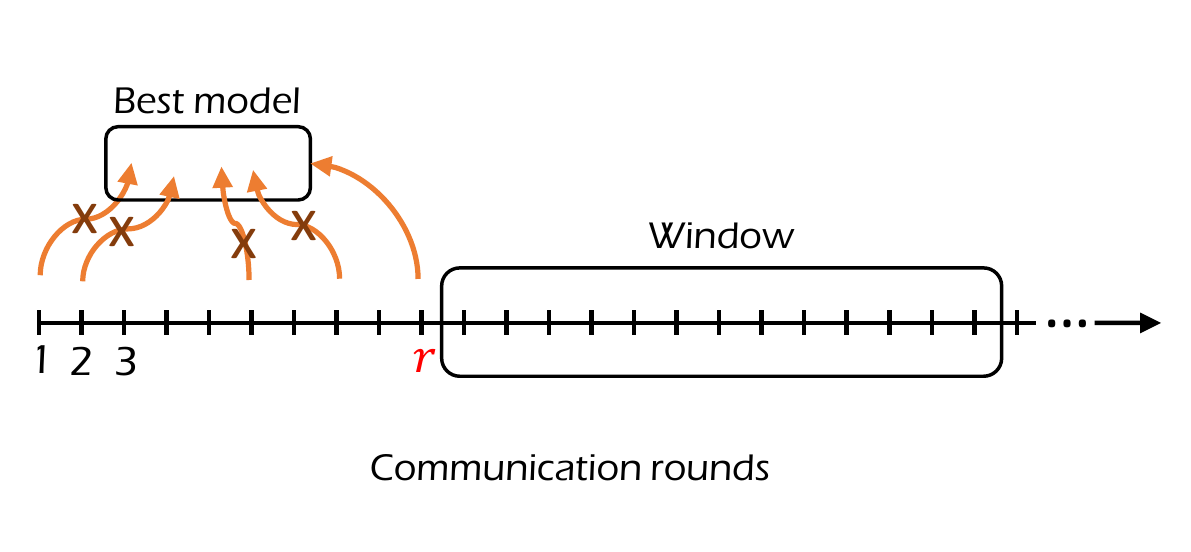}
\caption{Illustration of the delta-based early stopping patience algorithm; the current $r$ communication round has a better model that replaces the previous one, emptying the window. }
\label{Stop}
\end{figure}

The challenge of training an ML model in the federated setting is choosing the number of communication rounds. A large number of rounds incur unnecessary computations and waste of resources, leading to overfitting. Whereas a small number of rounds would result in a suboptimal model suffering underfitting. In this study, we also consider this challenge by developing a stopping algorithm that monitors the evaluation performance of the trained model at every iteration and halts the training process when no further improvement is expected. Instead of fixing the number of communication rounds, the delta-based early stopping algorithm sets a threshold (delta) for comparing the best model of the previous rounds with the currently trained model based on the validation dataset, as depicted in Fig. \ref{Stop}. If the current model has a better performance than the preceding best stored model, it will replace it; otherwise, the best model will remain. This comparison runs for several rounds defined as a window size; if the window is filled in without a better model being found, the training will stop and return the best model. We have implemented this algorithm on the FedAvg and FedTrees frameworks after an extensive study to determine the best delta and window size values. More details will be provided in the following section.

\section{Performance Evaluation and Results}\label{PE}
\subsection{Dataset Pre-Processing and Evaluation Methods}
\begin{figure}
\centering
\includegraphics[scale=0.4]{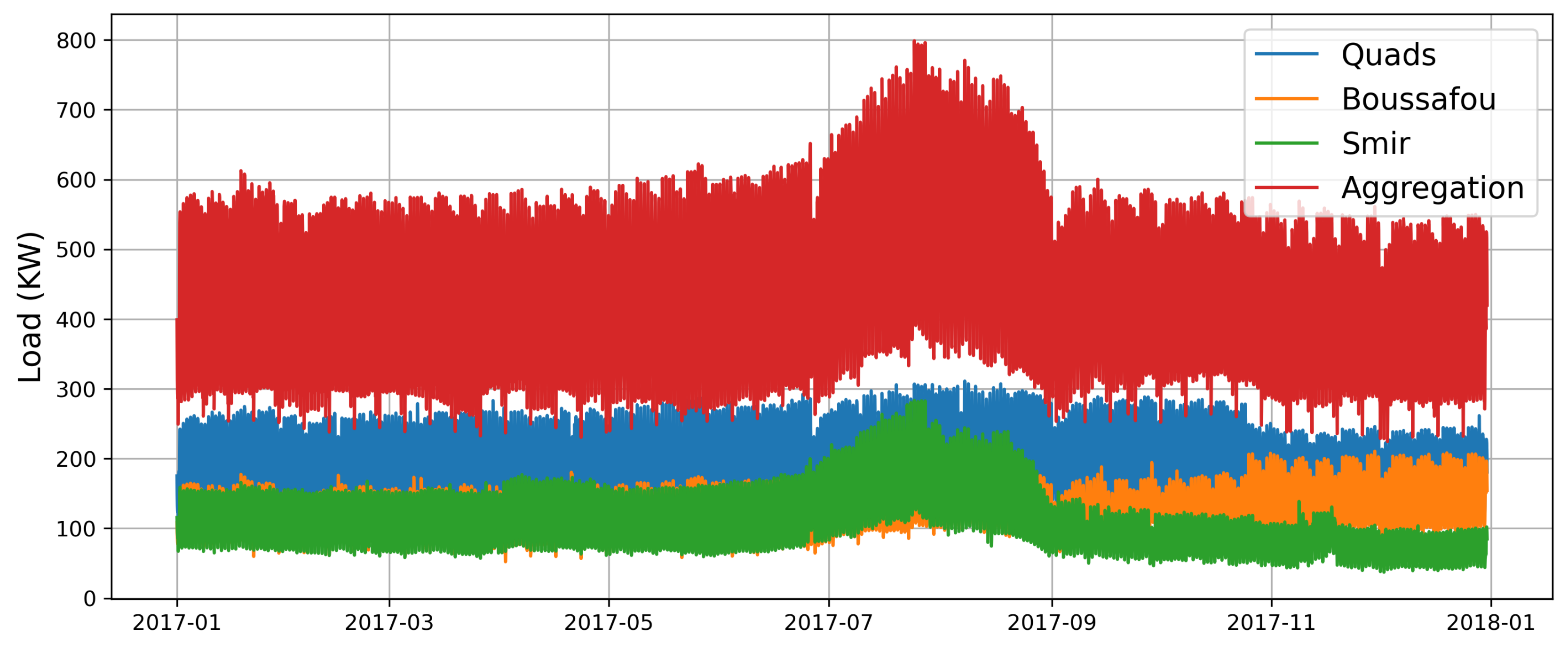}
\caption{Tetouan dataset preparation generated hourly power consumption of the three zones in addition to the aggregated power}
\label{Power}
\end{figure}

Since this study is intended to provide a practical framework capable of forecasting electricity load patterns in smart cities, it is essential to find an excellent dataset to evaluate the performance of FedTrees versus FedAvg. In this regard, the Tetouan power consumption dataset is selected for this purpose \cite{[36]}. This dataset was collected in 2017 at three different distribution substations from the zones: Quads, Boussafou, and Smir in Tetouan, a city located in north Morocco. In addition to providing power consumption information for every 10 minutes, the Tetouan dataset offers complementary data about the calendar and weather conditions. To prepare this dataset for our study, we initially converted the time scale from 10 minutes to 60 minutes because we are interested in predicting short-term loads for the next hour. Furthermore, we created two new dataset features; the aggregation feature that aggregates the power consumption of the three zones for use while performing centralised and distributed learning, and the previous hour aggregation (PrevHourAgg) feature that gives the aggregated feature reading of the previous hour. Fig. \ref{Power} demonstrates the hourly aggregated load as well as zones’ load information. Moreover, Table. \ref{features} shows the features that are considered to perform load forecasting.

\begin{table}
\centering
\caption{Tetouan dataset features used for load forecasting}
\begin{tabular}{c l|}
\toprule
\multicolumn{1}{c}{\textbf{Context}}    & \multicolumn{1}{c}{\textbf{Features}}                                                          \\ \midrule\midrule
\multicolumn{1}{|c|}{Calendar} & {\large \textcircled{\small 1} \small Month \large \textcircled{\small 2} \small Day \large \textcircled{\small 3} \small Hour}                      \\ \midrule
\multicolumn{1}{|c|}{Weather}  & {\begin{tabular}[c]{@{}l@{}}\large \textcircled{\small 4} \small Temperature \large \textcircled{\small 5} \small Humidity \large \textcircled{\small 6} \small Wind speed\\ \large \textcircled{\small 7} \small Diffuse flow \large \textcircled{\small 8} \small General diffuse flow\end{tabular}} 
    \\ \midrule
\multicolumn{1}{|c|}{Power}  & {\large \textcircled{\small 9} \small PrevHourAgg}                           \\ \bottomrule
\end{tabular}
\label{features}
\end{table}

Before commencing the training process, the default scale of the features is normalised using MinMax scaler into the range [0,1]. Accordingly, all features have the chance of contributing equally to model fitting and avoiding biasing. For evaluation purposes, we use the most popular metrics of mean absolute error (MAE) and mean absolute percentage error (MAPE), which are defined as \cite{[39]}:
\begin{equation}
    MAE=\frac{1}{n} \sum_{i=1}^{n}\left|y_{i}-\hat{y}_{i}\right|
\end{equation}

\begin{equation}
    MAPE=\frac{1}{n} \sum_{i=1}^{n}\left|\frac{y_{i}-\hat{y}_{i}}{y_{i}}\right|
\end{equation}


Where $y_{i}$ is the actual value, $\hat{y}_{i}$ is the predicted value, and $n$ represents the number of data samples.

\subsection{Simulation Setup}
To evaluate the performance of the FedTrees and FedAvg algorithms in power forecasting, we use the Persistence model as a benchmark, which is a naïve model that predicts the current value to be the same as the previous actual value. The forecasting problem is converted to a multivariate regression problem where we exploit various calendar, weather, and power features to predict power consumption for the next hour. For this purpose, we built LSTM and LGBM models based on the Random search strategy to discover the best combination of hyperparameters values. The LSTM model consists of a single hidden layer with 64 LSTM units that use the ReLU activation function and a dense output layer with one neuron. Also, it uses the Adam optimiser for compilation, the dataset is divided into 80/20\% train/test split ratio, batch size equals 30, and the number of epochs equals 300. Whereas the LGBM's boosting type, the number of trees\footnote{This hyperparameter is replaced by batch in FedTrees.}, max\_depth, learning rate, num\_leaves, and train/test split ratio are set to DART, 800, 12, 0.078, 30, and 80/20\%, respectively. Our simulation experiments are based on Python programs installed on a Windows operating system with Intel Xeon CPU E5-2620 @ 2GHz and 16GB RAM.

\subsection{Numerical Results}
We first investigate the performance of the selected models when performing the conventional approach of centralised training. Table \ref{Cen_Comp} demonstrates the results of the evaluation metrics as well as the required computation time for each model. The MAE and MAPE for the LSTM model are 0.02 and 3.04\%, respectively, while they are improved when using the LGBM model and reached 0.017 and 2.69\%, respectively. The fast-computing merit of the LGBM model is confirmed in this table which shows that it needs only two seconds to converge, while the LSTM model requires more than 97\% of computations compared to the LGBM model. 

\begin{table}[h]
\caption{Performance comparison between LSTM and LGBM models when performing centralised model training.}
\centering
\scalebox{0.9}{
\begin{tabular}{|l|l|l|l|}
\hline
\textbf{ML model} & \textbf{MAE} & \textbf{MAPE} & \textbf{Computation time} \\ \hline\hline
LSTM              & 0.019        & 3.04\%        & 77 seconds                \\ \hline
LGBM              & 0.017        & 2.69\%        & 2 seconds                 \\ \hline
\end{tabular}}
\label{Cen_Comp}
\end{table}

\begin{figure}
\centering
\includegraphics[scale=0.55]{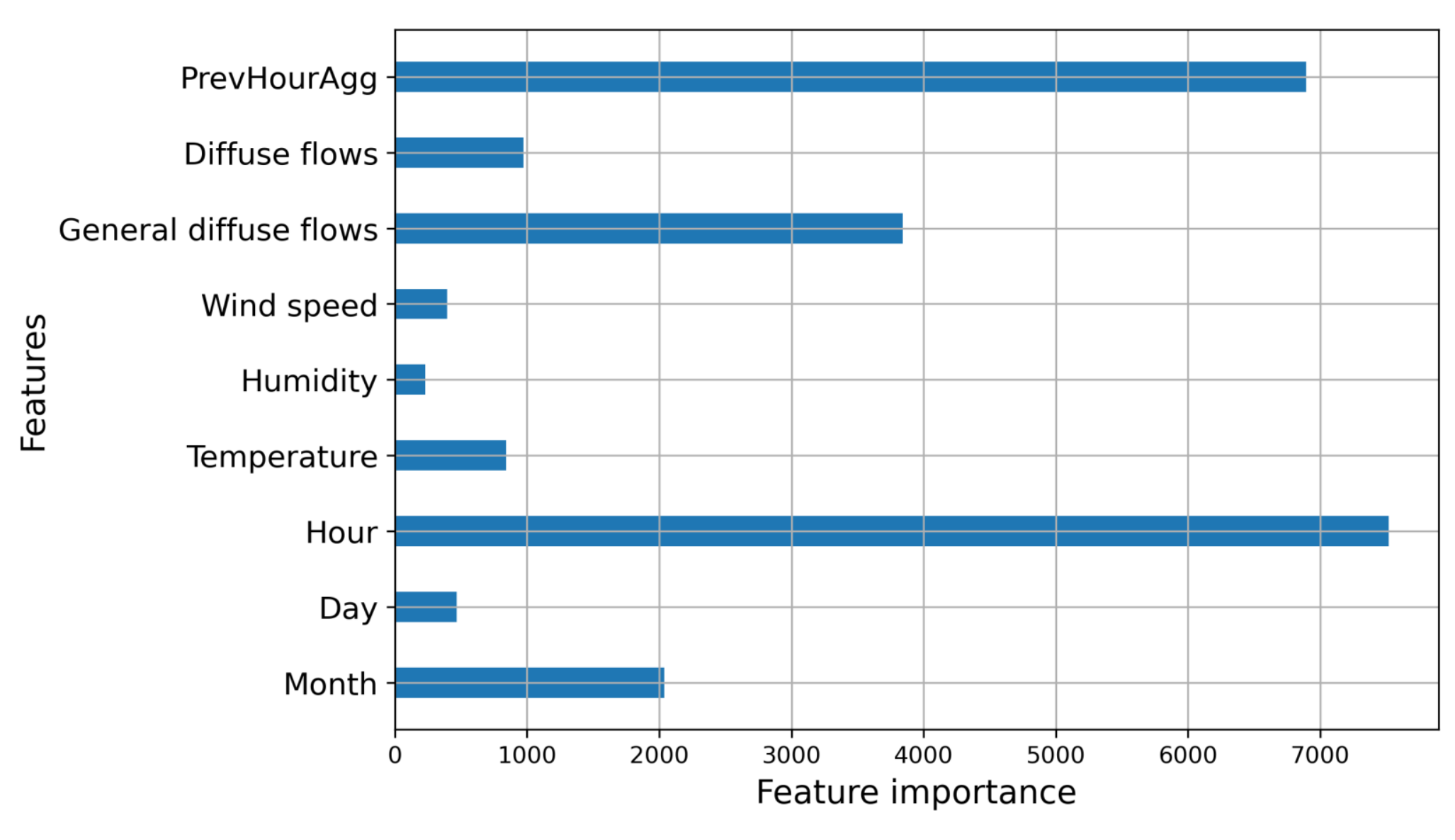}
\caption{The importance of each feature in forecasting power consumption.}
\label{FeatureImportance}
\end{figure}

\begin{figure}
\centering
\includegraphics[scale=0.4]{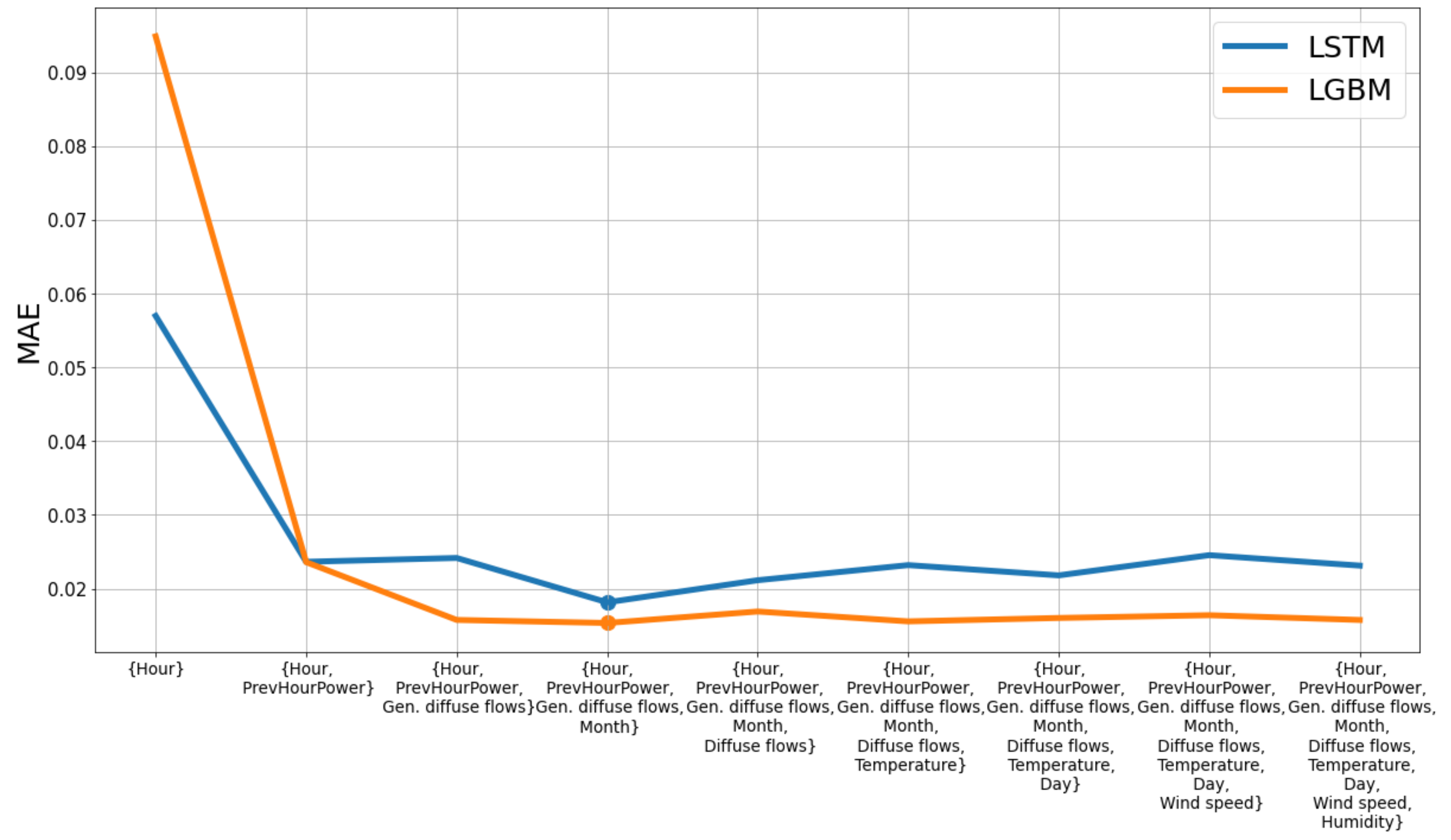}
\caption{MAE as a function of different features contributing in the prediction process.}
\label{FeaturesStudy}
\end{figure}

Furthermore, we conducted a study to explore the impact of every feature in predicting the target value. The LGBM model is equipped with a feature importance tool that can be used to find out what features contribute the most to the prediction of power values. Fig. \ref{FeatureImportance} demonstrates the results of the feature importance study from which we can conclude that Hour, PrevHourAgg, General diffuse flows, and Month have the most impact in predicting the power consumption values. Other features also contribute to the prediction but their contribution is less noticeable. Moreover, we performed a study to examine the prediction performance when using several numbers of features for the LSTM and LGBM models. Fig. \ref{FeaturesStudy} shows the best achieved MAE versus using different numbers of features which are ordered based on their rank in the feature importance study. This figure indicates that, in general, multivariate prediction gives improved performance over univariate; however, the best performance for both models is obtained when the four most important features are used.

Moving on to the FL setup, it is worth recalling that FedTrees is designed to accommodate cross-device and cross-silo settings. To the best of the authors' knowledge, FL has not been previously applied at the substation level in energy/power forecasting studies. Therefore, in this study, we focus on applying FedTrees and the FedAvg at the substation level. The FL setup for both algorithms comprises a central orchestrating server and three clients representing the three zones of the Tetouan dataset. The number of communication rounds is not fixed since we use the developed delta-based early stopping technique, discussed in Section \ref{Stopping}, to find the best round that yields the optimal trained model while alleviating the computation and communication costs. An extensive study is carried out to determine the best delta and window size values for the stopping algorithm, the findings of this study are given in Tables \ref{DWLSTM} and \ref{DWLGBM}. From Table \ref{DWLSTM}, the most computationally efficient and the best MAE/MAPE for the LSTM model are obtained when the delta and window size values are 0.00001 and 55, respectively. Regarding LGBM, Table \ref{DWLGBM} demonstrates that best values for the delta and window size are 0.00001 and 10, respectively. Table \ref{FinalFull} summarises the best results obtained for each of the Persistence, FedAvg, and FedTrees algorithms. The performance of the Persistence model is poor compared to other algorithms, and the FedAvg has the best performance, which is slightly better than that of FedTrees; however, FedTrees outperforms the FedAvg algorithm in terms of the communication rounds and the required computations. FedTrees only requires 65 rounds of communications that result in approximately 26 seconds of computations, while FedAvg requires a much higher number of communication rounds and computation time by a factor of 7.6 and 52.2, respectively, to achieve the same level of performance. 

\begin{table}
\centering
\caption{A study to determine the best values of delta and window size for LSTM-based FedAvg.}
\includegraphics[scale=0.48]{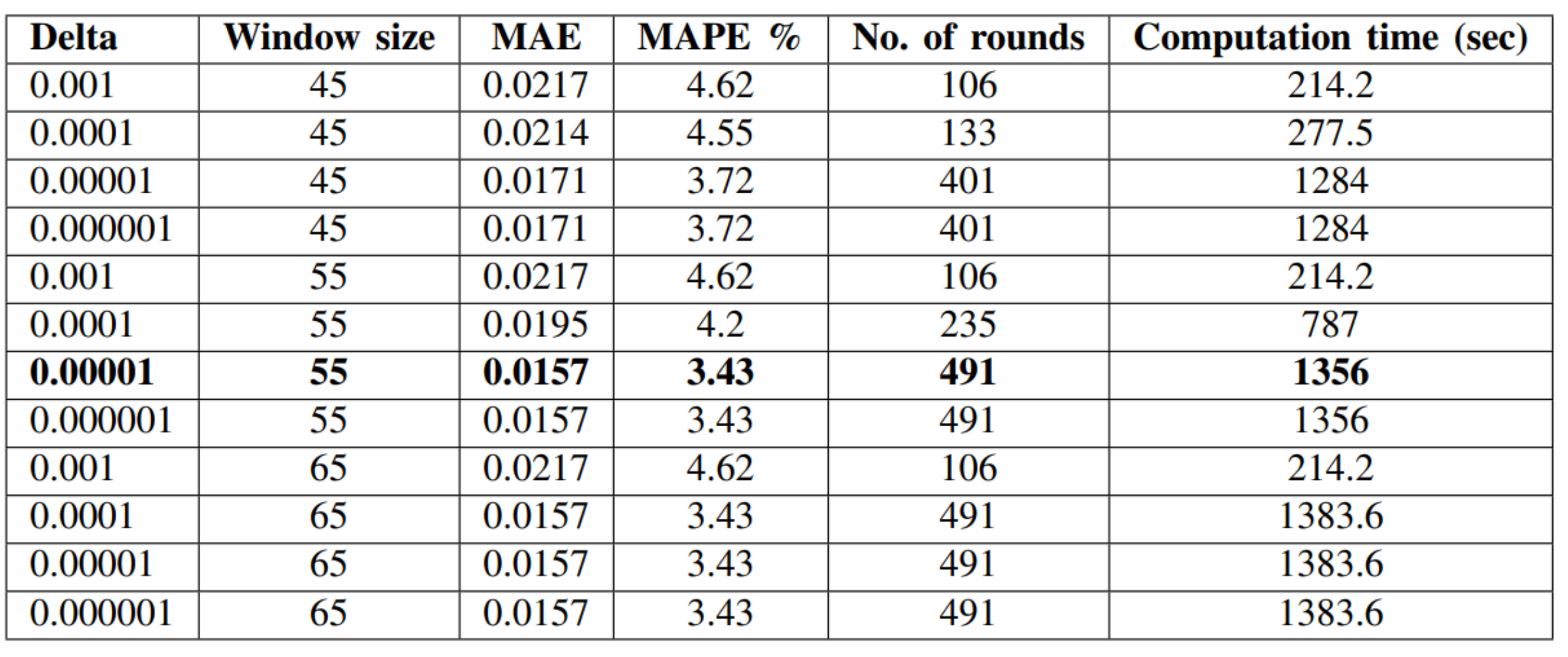}
\label{DWLSTM}
\end{table}

\begin{table}
\centering
\caption{A study to determine the best values of delta and window size for LGBM-based FedTrees.}
\includegraphics[scale=0.48]{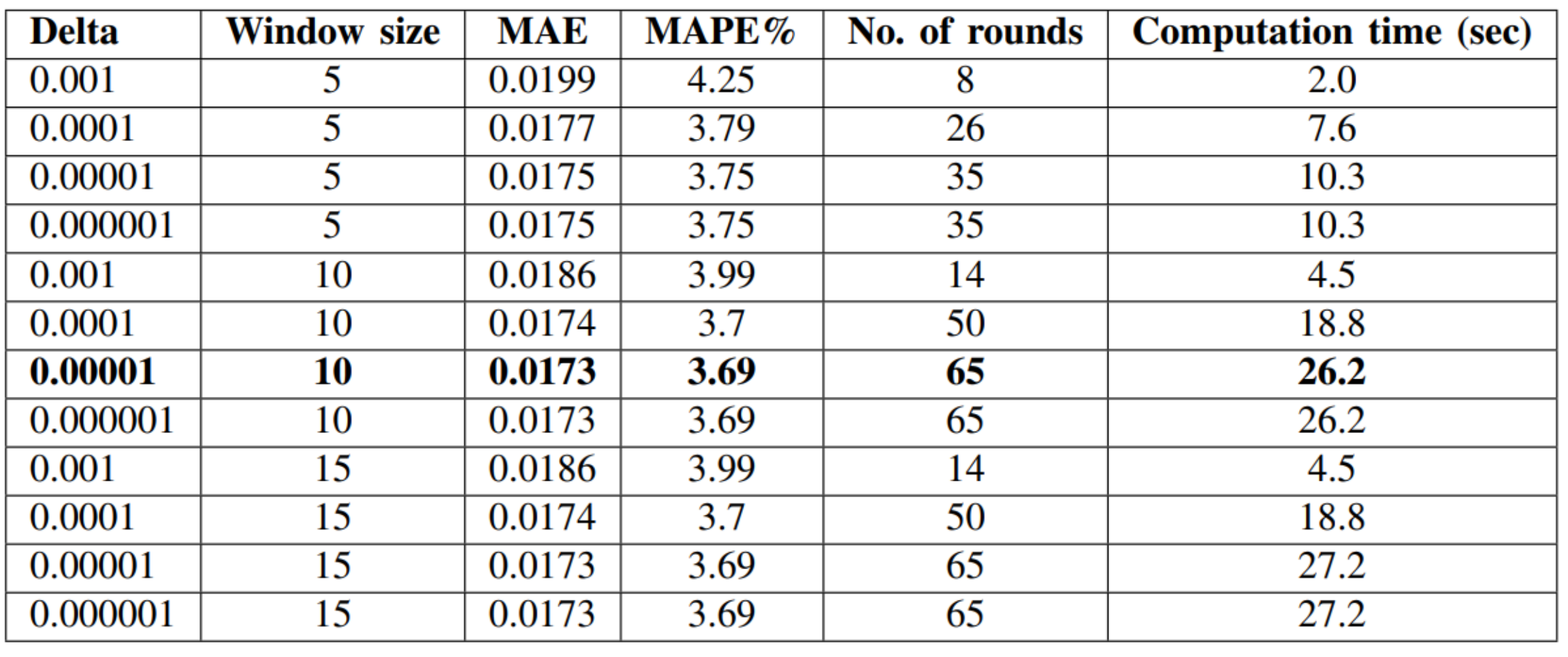}
\label{DWLGBM}
\end{table}

\begin{figure}
\centering
\includegraphics[scale=0.4]{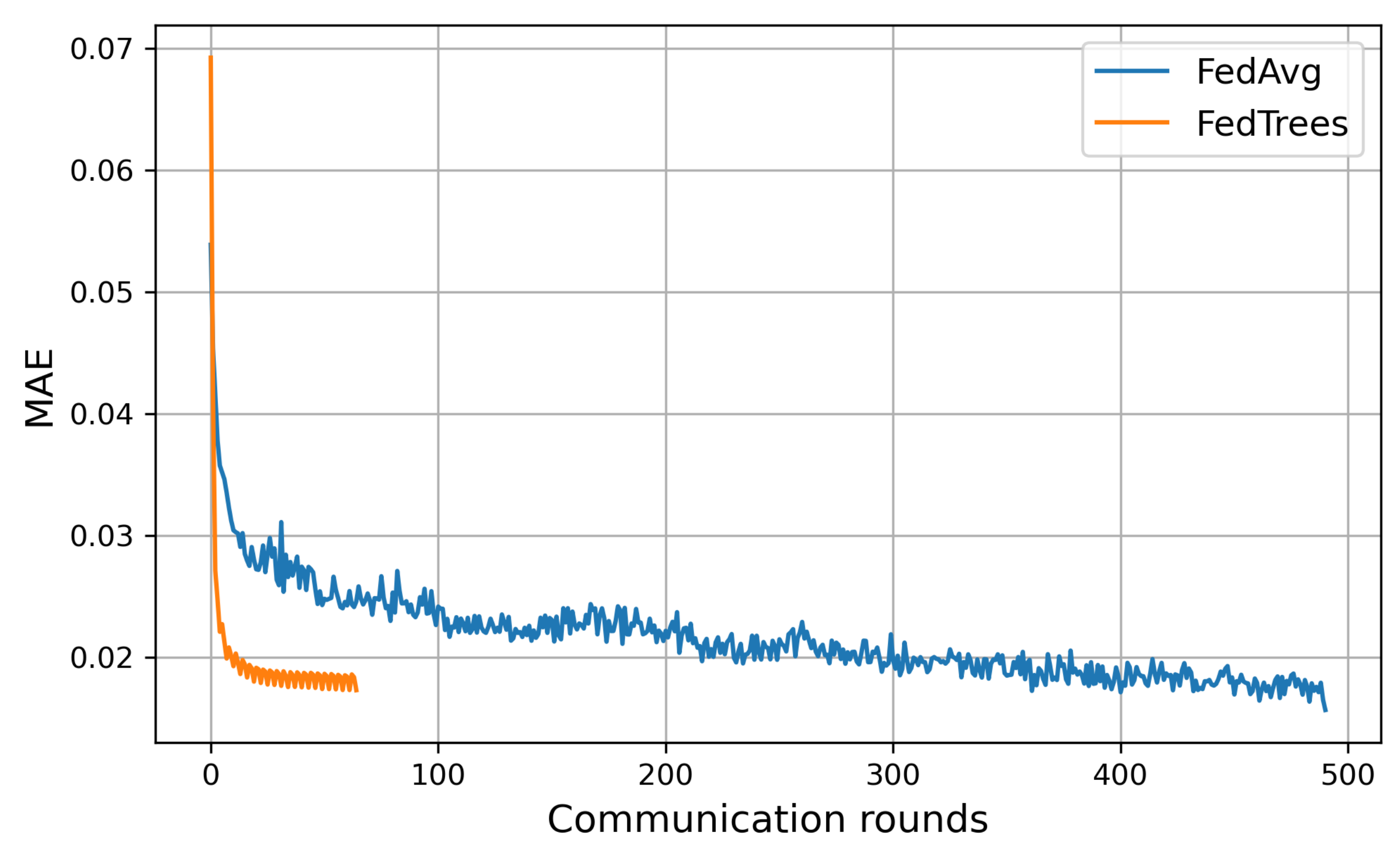}
\caption{MAE as a function of the communication rounds needed to train the global model of FedAvg and FedTrees.}
\label{CombinedMAE}
\end{figure}

\begin{figure}
\centering
\includegraphics[scale=0.32]{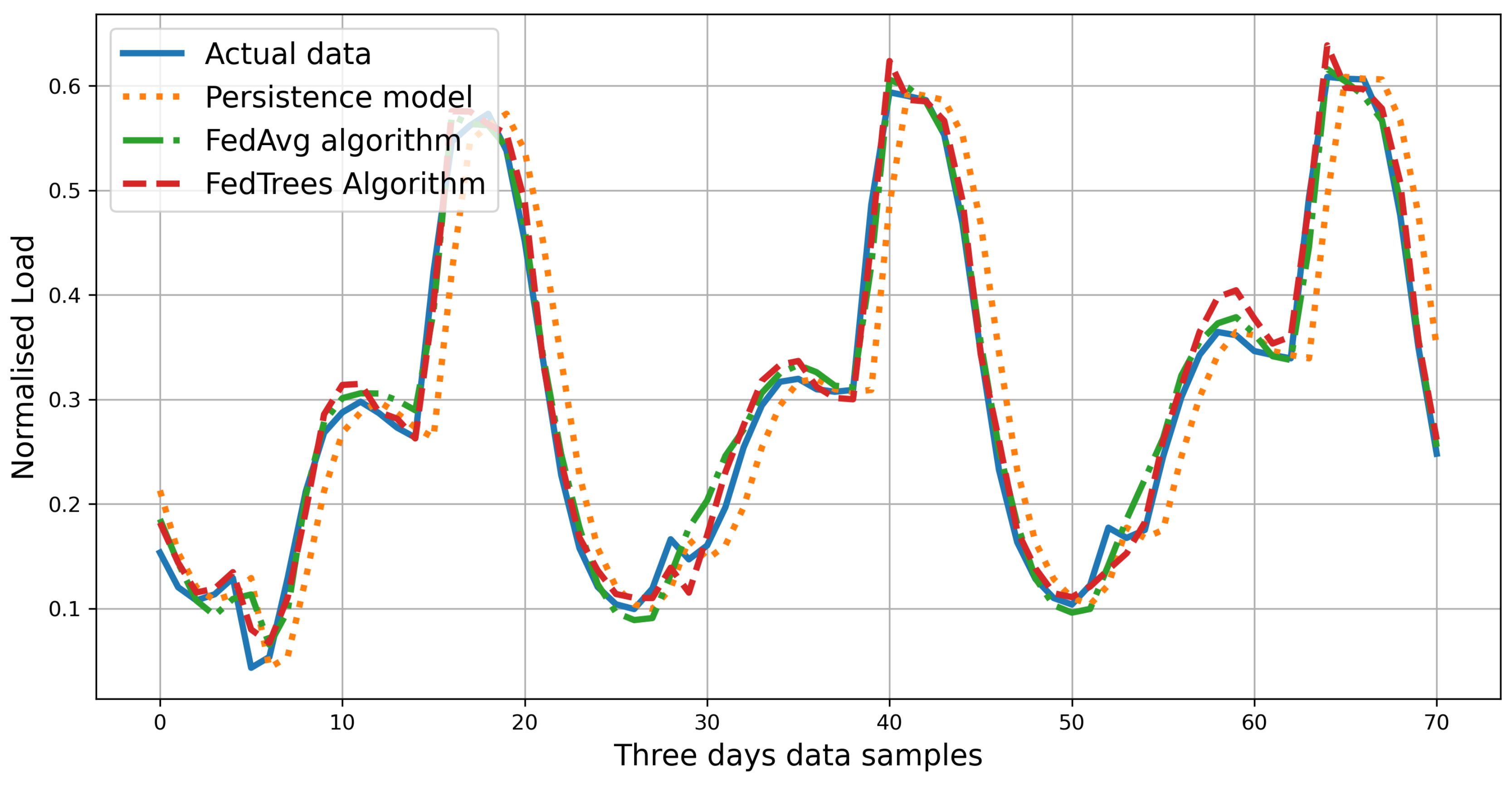}
\caption{Forecasting power consumption for three days.}
\label{CombinedForecast}
\end{figure}

Furthermore, Fig \ref{CombinedMAE} shows the convergence curve of MAE for both algorithms during the training of the global model. In addition, Fig \ref{CombinedForecast} shows the actual and the predicted power consumption from both algorithms and the baseline Persistence model. These figures indicate that FedTrees converges faster and achieves an outstanding performance compared to FedAvg.

\begin{figure}
\centering
\includegraphics[scale=0.4]{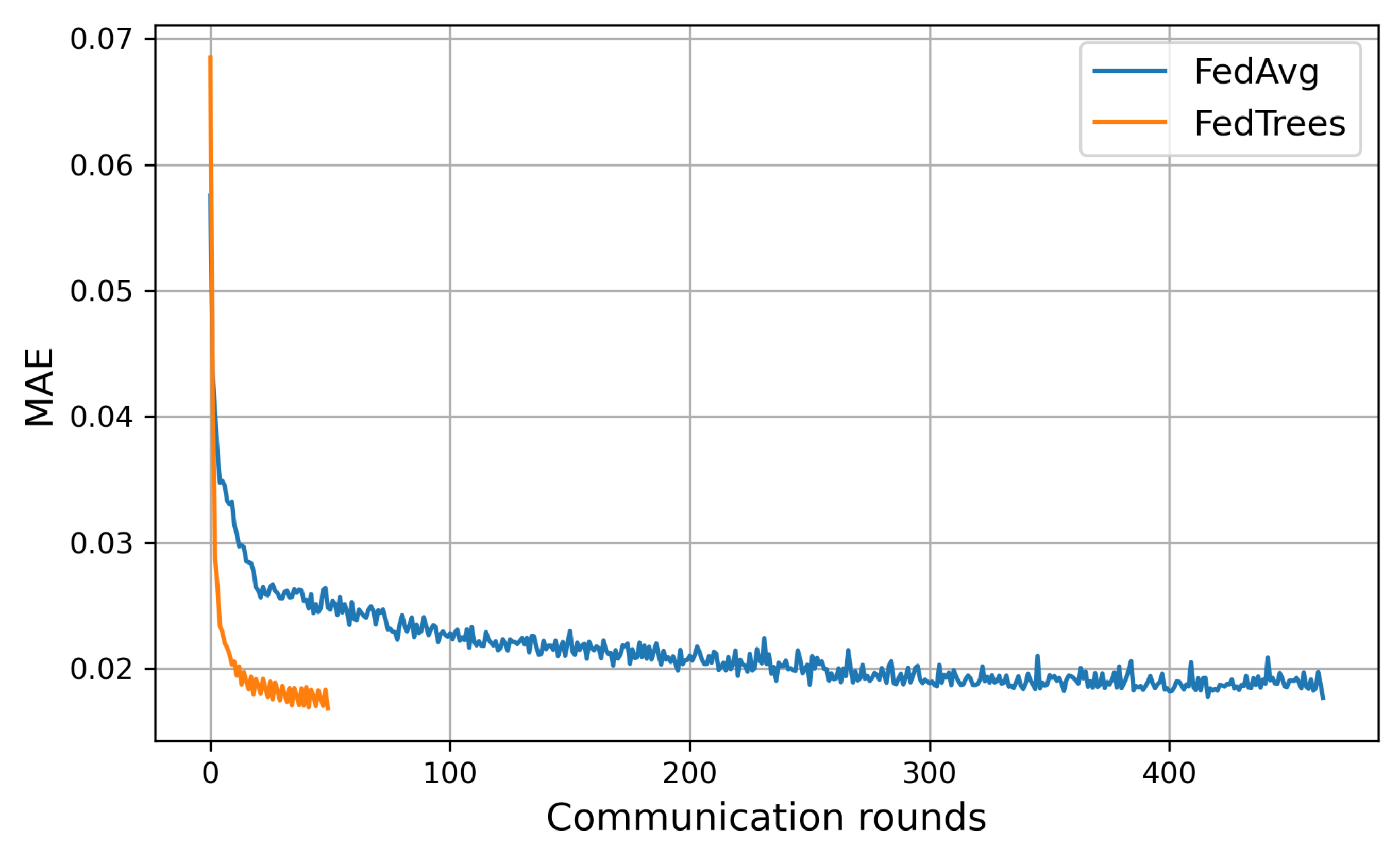}
\caption{MAE as a function of the communication rounds needed to train the global model of FedAvg and FedTrees when considering only the top four features.}
\label{CombinedMAE4F}
\end{figure}

\begin{figure}
\centering
\includegraphics[scale=0.32]{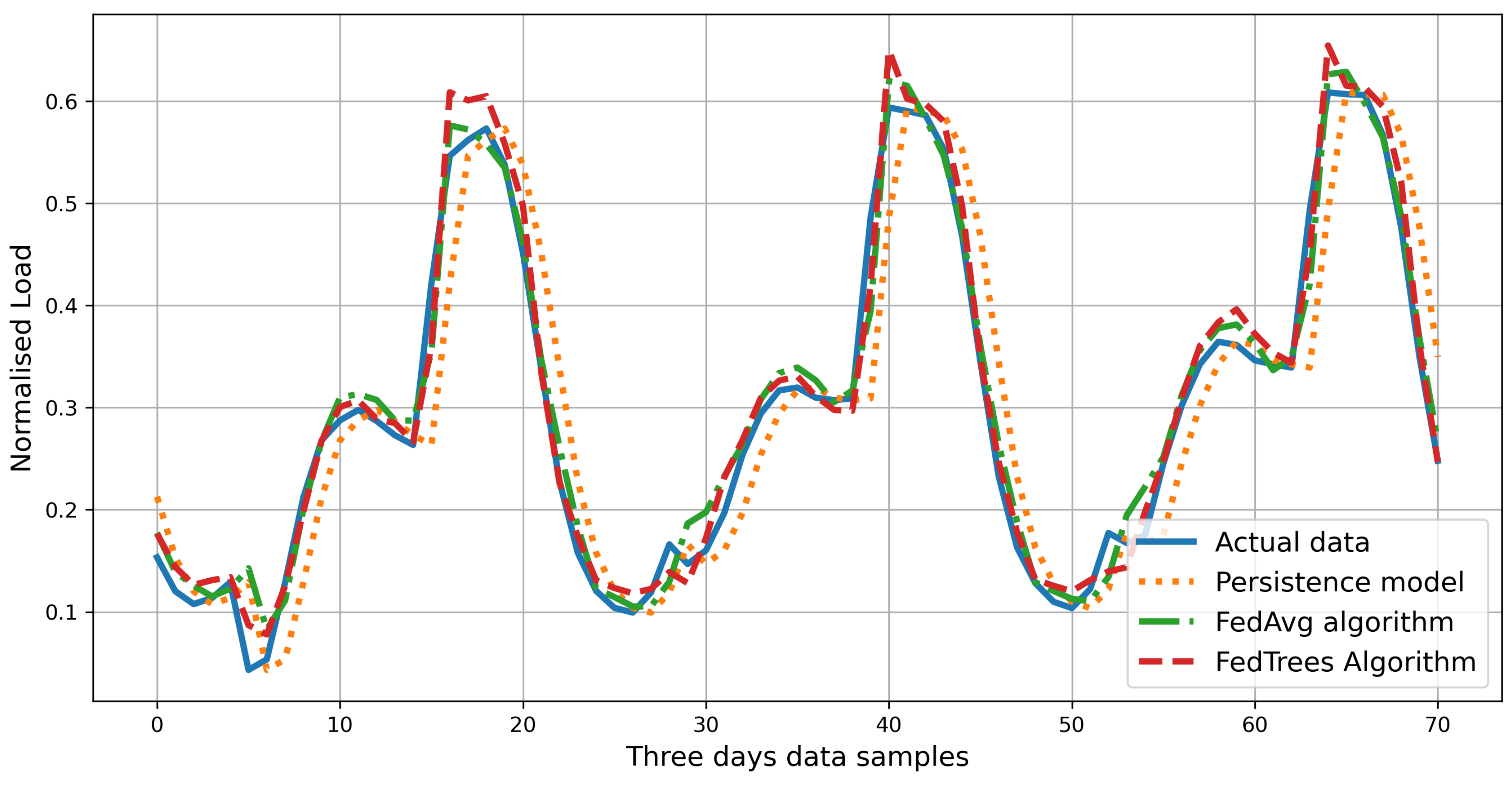}
\caption{Forecasting power consumption for three days when considering only the top four features.}
\label{CombinedForecast4F}
\end{figure}

Another study was conducted by looking only at the four most important features to see the impact of using fewer features on the required number of communication rounds, computation time, and model performance. Table \ref{Final4F} gives the outcomes of this study and indicates that the performance of FedTrees is improved when removing the less important features. However, this is not the case with FedAvg, as this table shows that its performance is slightly degraded compared to using all the features. The number of communication rounds is slightly less than that in the full feature study for both algorithms; however, this study also shows the outstanding performance of FedTrees as it requires far fewer rounds of communications and computation costs. Similarly, Fig. \ref{CombinedMAE4F} and \ref{CombinedForecast4F} give the MAE convergence curve and the actual and forecasted power consumption for both algorithms, respectively. Also, these figures ensure the superb overall performance of FedTrees. 

\begin{table}[h]
\centering
\caption{Performance results of the FedTrees compared to the FedAvg and the Persistence model. }
\scalebox{0.8}{
\begin{tabular}{|c|c|c|c|c|}
\hline
\textbf{Algorithm} & \textbf{MAE} & \textbf{MAPE\%} & \textbf{No. of rounds} & \textbf{Computation time} \\ \hline\hline
Persistence & 0.08   & 6.64 & N/A & N/A  \\ \hline
FedAvg      & 0.0157 & 3.43 & 491 & 1356 seconds \\ \hline
FedTrees    & 0.0173 & 3.69 & 65  & 26.2 seconds \\ \hline
\end{tabular}}
\label{FinalFull}
\end{table}

\begin{table}[]
\centering
\caption{Performance results of the FedTrees compared to the FedAvg and Persistence model when considering only the top four features. }
\scalebox{0.8}{
\begin{tabular}{|c|c|c|c|c|}
\hline
\textbf{Algorithm} & \textbf{MAE} & \textbf{MAPE\%} & \textbf{No. of rounds} & \textbf{Computation time} \\ \hline\hline
Persistence & 0.08   & 6.64 & N/A & N/A  \\ \hline
FedAvg      & 0.0177 & 3.93 & 465 & 1293 seconds \\ \hline
FedTrees    & 0.0168 & 3.54 & 50  & 8.8 seconds \\ \hline
\end{tabular}}
\label{Final4F}
\end{table}

\section{CONCLUSIONS} \label{Conclusion}
This paper aimed at developing FedTrees, a novel framework that incorporates ensemble learning, specifically the LGBM model, within the FL settings and coined to accommodate smart cities. Utilising the LGBM model transforms the FL into a highly efficient, fast processing, and scalable framework. Furthermore, instead of following the conventional fixed number of communication rounds method in FL, we developed a delta-based early stopping algorithm that monitors and stops the FL training process when no further enhancement is possible, thus ensuring achieving the desired training accuracy with the minimal use of computation and communication resources. FedTress is employed for the energy demand forecasting problem and benchmarked against LSTM-based FedAvg and Persistence model. Simulation results demonstrated that FedTrees has a remarkable performance in predicting short-term energy patterns and requires much less computation and communication than FedAvg with just 2\% and 13\%, respectively.

\bibliography{FL_Energy}

\end{document}